\pdfoutput=1

\documentclass[11pt]{article}

\usepackage[final]{acl}

\usepackage{times}
\usepackage{latexsym}

\usepackage{graphicx}
\usepackage{amsmath}
\usepackage{booktabs}
\usepackage{cleveref}
\usepackage{multirow}
\usepackage{hyperref}
\usepackage{amssymb}
\usepackage{pifont}
\usepackage{color, colortbl}
\usepackage{comment}
\usepackage{subcaption}
\usepackage{amsmath}
\usepackage{microtype}
\usepackage{bm}
\usepackage{caption}
\usepackage{enumitem}
\usepackage{xcolor}

\usepackage[T1]{fontenc}

\usepackage[utf8]{inputenc}

\usepackage{microtype}

\usepackage{inconsolata}

\usepackage{graphicx}

\title{Enhancing Multimodal Entity Linking with Jaccard Distance-based Conditional Contrastive Learning and Contextual Visual Augmentation}

\author{
    Cong-Duy Nguyen\textsuperscript{\rm 1}, \quad
    Xiaobao Wu\textsuperscript{\rm 1}\thanks{Corresponding Authors.}, \quad 
    Thong Nguyen\textsuperscript{\rm 2}, \quad 
    Shuai Zhao\textsuperscript{\rm 1}, \\
    \textbf{Khoi Le}\textsuperscript{\rm 3},  \quad 
    \textbf{Viet-Anh Nguyen}\textsuperscript{\rm 1},  \quad
    \textbf{Feng Yichao}\textsuperscript{\rm 1}, \quad
    \textbf{Anh Tuan Luu}\textsuperscript{\rm 1}$^*$ \\
  $^1$Nanyang Technological University, Singapore \\
  $^2$National University of Singapore, Singapore \quad $^3$VinAI Research, Vietnam \\
    \texttt{ \{nguyentr003,xiaobao.wu,shuai.zhao,yichao.feng,anhtuan.luu\}@ntu.edu.sg } \\
    \texttt{thongnguyen050999@gmail.com}
}

\begin{document}
\maketitle

\begin{abstract}

Previous research on multimodal entity linking (MEL) has primarily employed contrastive learning as the primary objective. However, using the rest of the batch as negative samples without careful consideration, these studies risk leveraging ``easy'' features and potentially overlook essential details that make entities unique. In this work, we propose JD-CCL (Jaccard Distance-based Conditional Contrastive Learning), a novel approach designed to enhance the ability to match multimodal entity linking models. JD-CCL leverages meta-information to select negative samples with similar attributes, making the linking task more challenging and robust. Additionally, to address the limitations caused by the variations within the visual modality among mentions and entities, we introduce a novel method, CVaCPT (Contextual Visual-aid Controllable Patch Transform). It enhances visual representations by incorporating multi-view synthetic images and contextual textual representations to scale and shift patch representations. Experimental results on benchmark MEL datasets demonstrate the strong effectiveness of our approach.

\end{abstract}

\section{Introduction}

\begin{figure}[t]
    \centering
    \includegraphics[width=0.45\textwidth]{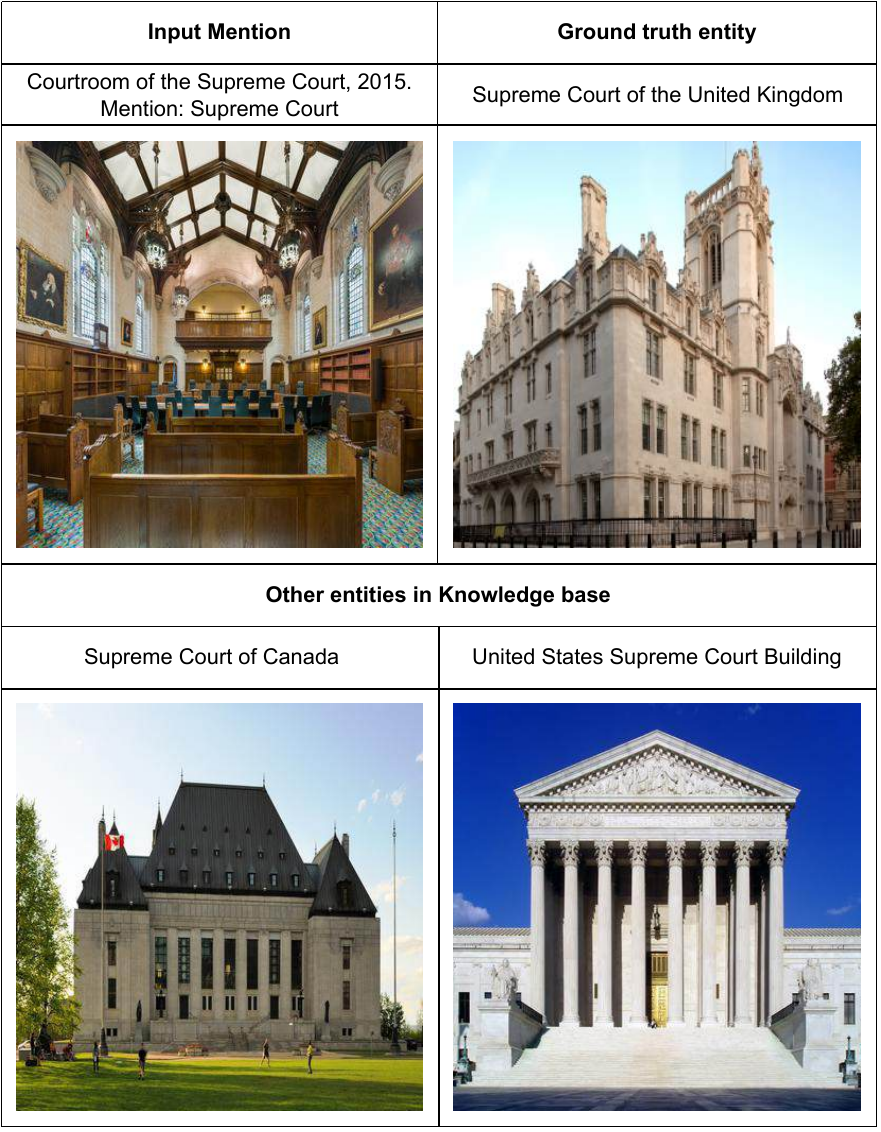}
    \caption{Example from WikiDiverse: The first two samples are the sentence input and the ground truth entity for its mention. The last two are other "Supreme Court" entities taken from the knowledge base.}
    \label{fig:ex1}
    \vspace{-10pt}
\end{figure}

\begin{figure*}[t]
    \centering
    \includegraphics[width=0.92\textwidth]{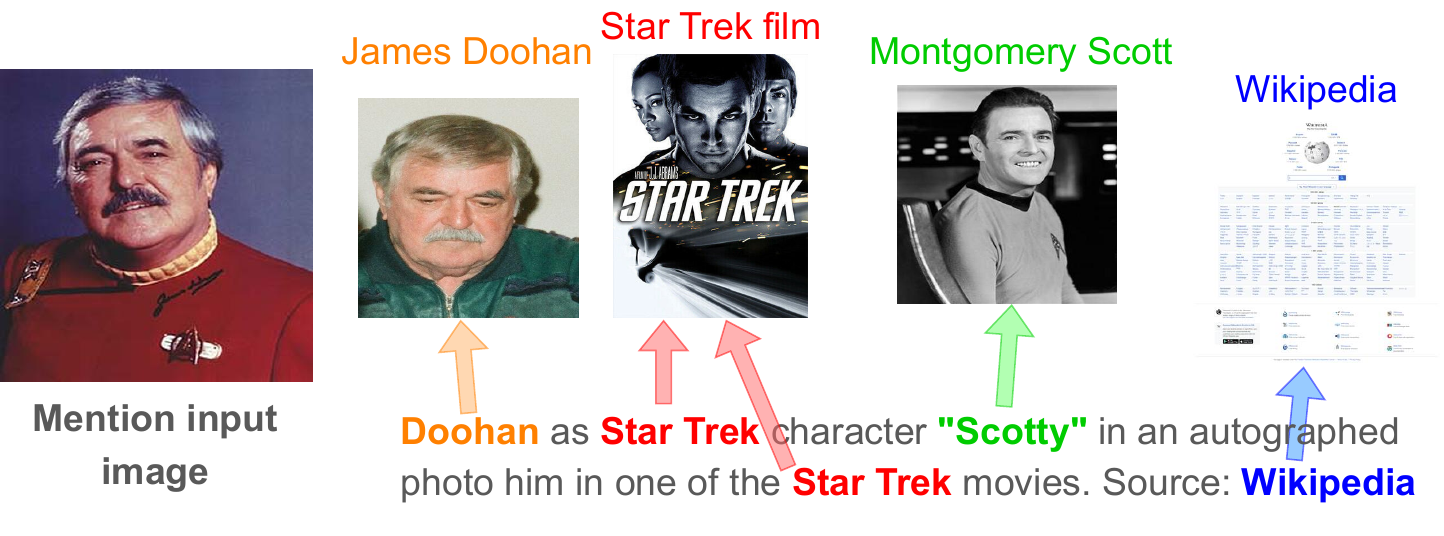} 
    \caption{Illustration of challenge involving disparities within the visual modality. A sample from WikiDiverse includes the mention sentence and its image talking about the actor Doohan. Input sentence contains multiple entities from the knowledge base: Doohan, Star Trek, Scotty, Wikipedia.}
    \label{fig:ex2}
    \vspace{-10pt}
\end{figure*}

\begin{figure}[t]
    \centering
    \includegraphics[width=\linewidth]{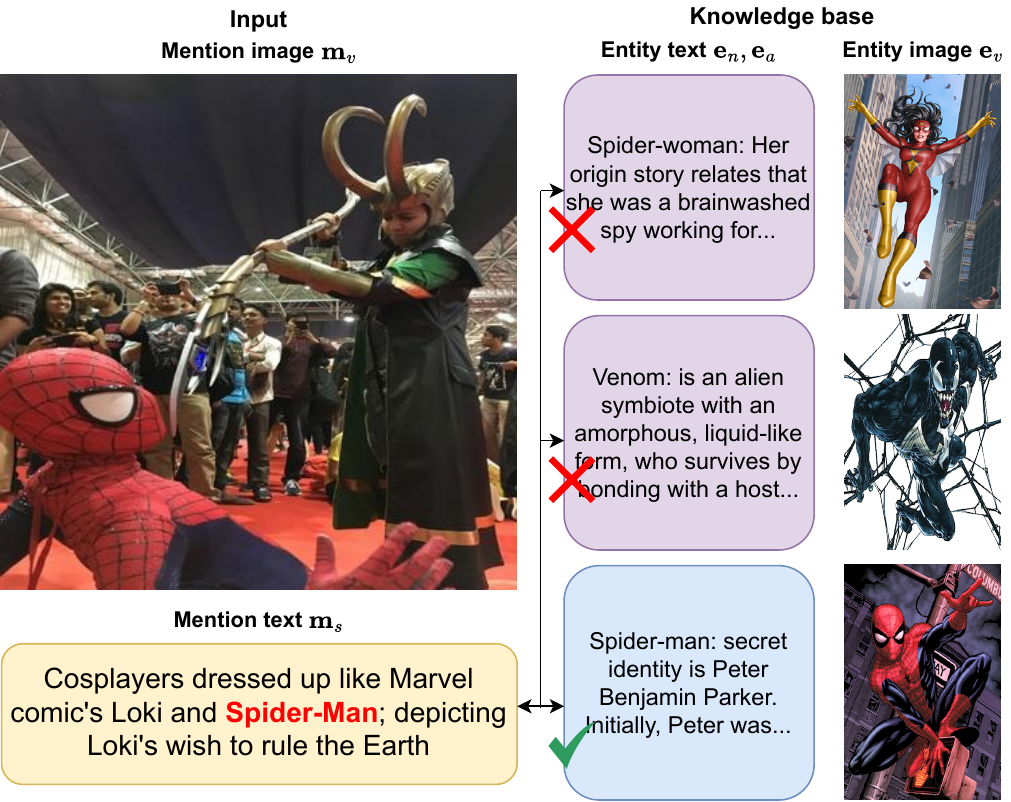}
    \caption{Task definition of Multimodal Entity Linking.}
    \label{fig:task}
    \vspace{-10pt}
\end{figure}

Multimodal Entity Linking (MEL)~\cite{song2023dualway,pmlr-v216-yang23d,shi2024generative} aims to connect visual and textual references to the entities in a multimodal knowledge graph~\cite{song2023dualway,pmlr-v216-yang23d,shi2024generative}.
As exemplified in \Cref{fig:task}, MEL connects the Spider-Man referecnes in the image and input sentence to the Spider-Man entity in the multimodal knowledge graph.
This alleviates textual ambiguities by incorporating visual cues such as character portraits,
which benefits various downstream applications, like information extraction~\cite{yaghoobzadeh-etal-2017-noise}, question answering~\cite{DBLP:conf/emnlp/LongprePCRD021}, and search engines~\cite{DBLP:conf/sigir/GerritseHV22}.


However, current MEL methods are limited by their random negative sampling and understudy the potential issue of varied visual representation.
First, these methods tend to employ contrastive learning to align input sentences and entities, but limited by its random negative sampling~\cite{DBLP:conf/acl/WangTGLWYCX22,luo2023multigrained}.
They generally randomly draw negative samples from the datasets.
This may lead to unfair predictions due to numerous similar entities in the knowledge base,
for example, different but highly similar ``Supreme Court'' entities in \ref{fig:ex1}. 
Using them as negative samples hinders the connection between input sentences and correct entities.
Second, existing MEL approaches overlooked the dissimilar visual modality between mention and target entity.
For instance, \ref{fig:ex2} shows that the input image refers to an actor ``Doohan'', but the input sentence includes four distinct entities.
As a result, this greatly disturbs linking the input image to the entities in the knowledge base. 
For further specificity, the "Doohan" image on the left illustrates the sentence, offering significant utility to the model for user queries related to \textcolor{orange}{"Doohan"} within the knowledge base due to image-text similarity. However, it could potentially introduce interference if the user intends to query \textcolor{blue}{"Wikipedia"}. Hence, pre-linking control of the visual modality, contingent upon the mentioned target, emerges as a crucial consideration in the Multimodal Entity Linking (MEL) task, particularly when other modalities coexist alongside textual input.

To address the limitation of random negative sampling, we present a Jaccard Distance-based conditional contrastive learning (JD-CCL) approach for MEL. It leverages meta-information to mitigate the influence of prominent attributes during contrastive pre-training.
JD-CCL automatically utilizes the information provided by each entity in the knowledge base to sample hard negative entities with nearly identical attributes. This strategy makes it more challenging for the model to link the mention with the correct entity, as it cannot rely on simple attributes (such as human, building, or animal) but must distinguish positive pairs from negative ones based on more complex attributes. To identify similar entities, we define a pre-processing stage that sorts the Jaccard similarity scores between the meta-attributes between entities in knowledge base.
This process brings more difficult and nuanced comparisons during training,
which improves the linking ability by diminishing the impact of ``easy'' attributes in contrastive training.

Moreover, to address the discrepancy challenge of visual modality in image-text pairs, we introduce the Contextual Visual-Aid Controllable Patch Transform (CVa-CPT) module.
By leveraging synthetic images from a text-to-image diffusion, the model can better control essential features of the input image based on specific mention. 
Consequently, the model can selectively control useful patch representations from the input image 
rather than using all image patches with the same contribution. Additionally, for sentences containing multiple entities, each entity will have a unique input image, enhancing diverse visual representation according to the entity being queried.

Our contributions can be summarized as follows:
\begin{itemize}[leftmargin=*,itemsep=0pt]
\item We propose a novel Jaccard Distance-based Conditional Contrastive Learning (JD-CCL). This method leverages meta-attributes to draw conditional negative samples with similar features during training, enabling the model to identify exact characteristics for decision-making.

\item We introduce the Contextual Visual-Aid Controllable Patch Transform (CVa-CPT) module, which enhances visual modality representations with synthetic images and contextual information from the mention sentence.

\item We evaluate our approach on three benchmark datasets: WikiDiverse, RichpediaMEL, and WikiMEL, and show that our method outperforms previous state-of-the-art approaches.
\end{itemize}

\section{Related Work}

\paragraph{Multimodal Entity Linking} 

In the landscape of social media and news, where content frequently encompasses both text and visuals, integrating textual and visual information for entity linking is essential and pragmatic. In an early pioneering effort, \cite{DBLP:conf/acl/CarvalhoMN18} utilized images to enhance entity linking in response to ambiguous and incomplete references in social media posts. Advancing this, \cite{10.1145/3477495.3531867} investigated the inter-modal interactions through a cross-attention mechanism between text and vision, integrating a gated hierarchical architecture. To mitigate the impact of noisy and irrelevant images, \cite{DBLP:conf/dasfaa/ZhangLY21} assessed the relevance of images by correlating their category with the semantic content of text mentions, filtering images using a predefined threshold. \cite{DBLP:conf/mm/GanLWWHH21} compiled a dataset featuring extensive movie reviews, associated entities, and images. More recently, research~\cite{DBLP:journals/dint/ZhengWWQ22} has included the use of scene graphs from images to achieve object-level encoding, thus enhancing the granularity of visual semantic cues. MIMIC~\cite{luo2023multigrained} presents a multi-grained multimodal interaction network designed for the multimodal entity linking task. 
GMEL~\cite{shi2024generative} offers a Generative Multimodal Entity Linking framework employing LLMs that directly produces names of target entities.
MMELL~\cite{pmlr-v216-yang23d} introduces a module that jointly extracts features to learn representations from both visual and textual inputs for context and entity candidates. DWE~\cite{song2023dualway} describes a dual-way enhancement framework where the query benefits from refined multimodal information, and Wikipedia is used to augment the semantic representation of entities.

\paragraph{Contrastive learning}

Contrastive Learning \cite{1467314, nguyen2021contrastive, nguyen2025meta, nguyen2024multi, nguyen2024motion, nguyen2023demaformer} has become increasingly prominent in various fields \cite{nguyen2022adaptive,nguyen2023improving,nguyen2024topic,Wu2020short,wu2022mitigating,wu2023infoctm,wu2023effective,wu2024dynamic,wu2024traco,wu2024topmost,wei2024learning}. Established training methods such as N-Pair Loss \cite{NIPS2016_6b180037}, Triplet Margin Loss \cite{Balntas2016LearningLF}, and ArcCon \cite{zhang-etal-2022-contrastive} are foundational in metric learning. In supervised learning scenarios, key methodologies include Center loss \cite{10.1007/978-3-319-46478-7_31}, SphereFace \cite{8100196}, CosFace \cite{DBLP:journals/corr/abs-1801-09414}, and ArcFace \cite{DBLP:journals/corr/abs-1801-07698}, which are extensively applied in computer vision and natural language processing. Recent developments in contrastive learning incorporate additional conditional variables to enhance representation quality, involving auxiliary attributes \citep{tsai2021integrating}, information pertinent to the downstream task \citep{tian2020makes}, downstream labels \citep{khosla2020supervised, kang2020contragan}, or data embeddings \citep{tsai2022conditional, wu2020conditional}. These advancements allow for the extension of contrastive self-supervised learning into weakly supervised \citep{tsai2021integrating}, semi-supervised \citep{tian2020makes}, or fully supervised frameworks \citep{khosla2020supervised}. The work by \citep{khosla2020supervised} specifically seeks to enhance fairness, rather than solely focusing on representation quality, by employing sensitive attributes from the dataset as the conditional variables in a self-supervised context.

\section{Problem Formulation}

A multimodal knowledge base includes a set of entities $\mathrm{E}= \left \{\mathbf{E}_i \right \}_{i=1}^{N}$.
Each entity is denoted as $\mathbf{E}_i=(\mathbf{e}_{n_i}, \mathbf{e}_{v_i}, \mathbf{e}_{d_i}, \mathbf{e}_{a_i})$ 
, represent entity name, entity images, entity description, and entity attributes, respectively. In this representation, the $\mathbf{E}_i$ components correspond to the entity's name, images, description, and attributes, respectively.
As our study emphasizes linking local-level entities, textual input is formatted as sentences rather than as documents.
Specifically, a mention and its context are represented as $\mathbf{M}_j=(\mathbf{m}_{w_j}, \mathbf{m}_{s_j}, \mathbf{m}_{v_j})$, where $\mathbf{m}_{w_j}, \mathbf{m}_{s_j}$, and $\mathbf{m}_{v_j}$ denote the words of the mention, the sentence containing the mention and the associated image, respectively. The related entity of the mention $\mathbf{M}_j$ in the knowledge base is $\mathbf{E}_i$. The MEL task is illustrated in Figure~\ref{fig:task}.

\begin{figure}[t]
    \centering
    \includegraphics[width=\linewidth]{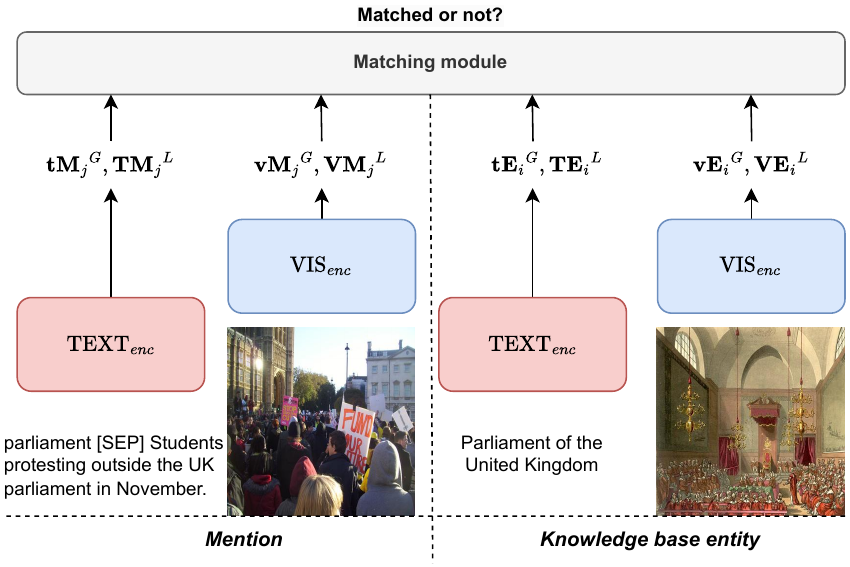} 
    \caption{The overall architecture of MEL problem.}
    \label{fig:overview}
    \vspace{-7pt}
\end{figure}

\section{Methodology}

In this section, we first introduce the overall framework and then go through the details of our approaches. We visualize the overall architecture in Figure~\ref{fig:overview}.

\subsection{Overall Architecture}

\paragraph{Visual Encoding}

Following previous work, we utilize the pre-trained Vision Transformer (ViT)~\cite{DBLP:conf/iclr/DosovitskiyB0WZ21} as our visual backbone to encode image features, denoted as $\mathrm{VIS}_{enc}$. Each entity image $\mathbf{e}_{v_i}$ (or mention image $\mathbf{m}_{v_j}$) is resized to $C \times H \times W$ pixels and divided into flattened 2D patches of size $n = H \times W /P^2$. These patches undergo projection and multi-layer transformer processes. Subsequently, a fully connected layer adjusts the output hidden states. The resulting hidden states, denoted as $\mathbf{V}{\mathbf{E}_i}$, include a global feature $\mathbf{v}{\mathbf{E}_i}^G$ represented by the special token \texttt{[CLS]}, and local features $\mathbf{V}{\mathbf{E}_i}^L$ (similarly for mention $\mathbf{v}{\mathbf{M}_j}^G$ and $\mathbf{V}{\mathbf{M}_j}^L$).

\paragraph{Textual Encoding}

We employ a Transformer-based architecture as the textual encoder (BERT~\cite{DBLP:conf/naacl/DevlinCLT19}), denoted as $\mathrm{TEXT}_{enc}$. For mentions $\textbf{M}_j$, we concatenate the mention words with the sentence containing the mention to create the input sequence. We obtain $\mathbf{t}{\mathbf{M}_j}^G$ and $\mathbf{T}{\mathbf{M}_j}^L$ as the global and local textual features of mention $\mathbf{M}_j \in \mathbb{R}^{(l_e + 1) \times d_t}$, respectively. Here, $d_T$ is the dimension of textual output features, and $l_e$ is the length. For entities $\textbf{E}_i$, entity inputs are formed by concatenating the entity name with its attributes, denoted as $\mathbf{e}_{a_i}$. Following the mentioned processing, the resulting hidden states $\mathbf{T{\mathbf{E}_i}}$ are structured as $\left [ \mathbf{t}{\text{[CLS]}}^0; \mathbf{t}{\mathbf{E}_i}^1; \dots; \mathbf{t}{\mathbf{E}_i}^{l_e} \right ] \in \mathbb{R}^{(l_e + 1) \times d_t}$.


\paragraph{Matcher}

In this step, the model calculates the similarity matching score between input mention text-image pairs and knowledge entities to achieve correct matching. For instance, using the CLIP~\cite{DBLP:conf/icml/RadfordKHRGASAM21} template, we calculate the cosine similarity score between the global representation (\text{[CLS]} token) the mention input with the entity. Alternatively, MIMIC~\cite{luo2023multigrained} introduced the Multi-Grained Multimodal Interaction Network, which captures and integrates the fine-grained representation of global and local cues of each modality. Overall, the matcher acts as a scoring function $\hat{y} = \mathrm{F}(\mathbf{M},\mathbf{E})$ between the mention $\mathbf{M}$ and the entity $\mathbf{E}$.

\paragraph{Objective}

Based on the score calculated above, we jointly train both the encoding layer and the matcher using a contrastive training loss function. This approach allows the model to learn to assign higher scores to positive mention-entity pairs and lower scores to negative mention-entity pairs. This loss function can be formulated as follows.

\vspace{-5pt}

\begin{equation}\resizebox{.95\hsize}{!}{$
    \mathrm{L} = -\log \frac{\exp \left(\mathrm{F}(\mathbf{M}, \mathbf{E}_{pos}) \right)}{\exp \left(\mathrm{F}(\mathbf{M}, \mathbf{E}_{pos}) \right )+ \sum_i \exp \left( \mathrm{F}(\mathbf{M}, \mathbf{E}_{neg,i}) \right)}
    \vspace{-7pt} $}
\end{equation}
\subsection{Conditional Contrastive learning}

\begin{figure*}[t]
\centering
\includegraphics[width=0.92\textwidth]{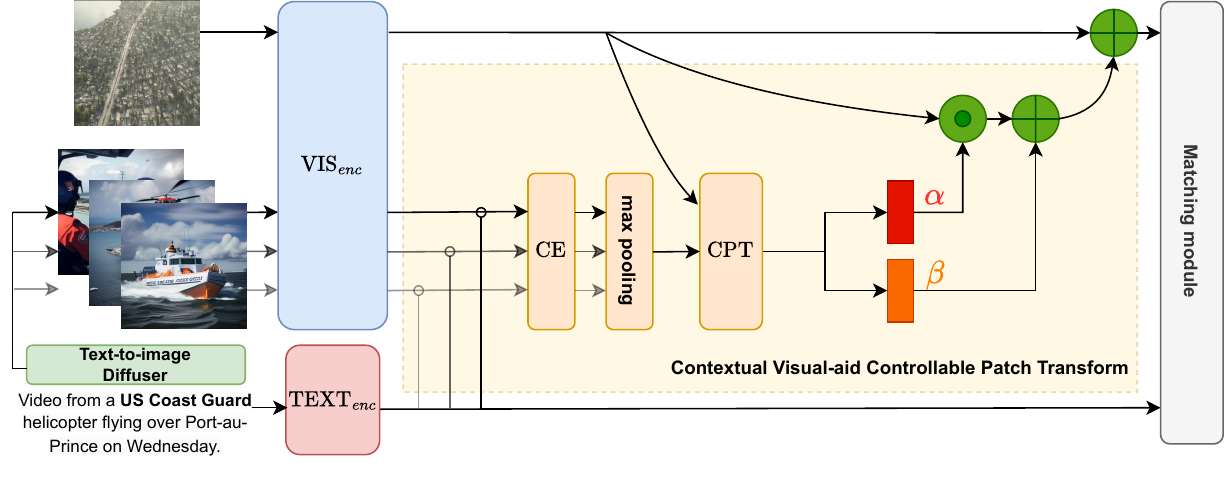} 
\caption{The overall architecture of Contextual Visual-aid Controllable Patch Transformation. The model incorporates a text-to-image diffuser, visual encoder, and text encoder, followed by a contextual visual-aid controllable patch transform (CVaCPT). The generated images and input sentence will help the model selectively control visual representation through the CVaCPT, including Contextual encoder, Max pooling, and Controllable patch transform.}
\label{fig:framework}
\vspace{-10pt}
\end{figure*}

Conditional Contrastive Learning (CCL)~\cite{ma2022conditional} was proposed to enhance the fairness of contrastive self-supervised learning (SSL) methods. CCL mitigates the influence of sensitive attributes during contrastive pre-training by sampling positive and negative pairs that share the same attribute (e.g., gender, race). The CCL approach reduces the impact of the sensitive attribute $Z$ by treating $Z$ as the conditional variable between $X$ and $Y$, focusing on scenarios where $Z$ is readily available in the dataset (e.g., gender, race, or age). The proposed \textbf{Conditional Contrastive Learning} objective is designed to address this effectively.

\vspace{-5pt}
\begin{equation}
\small
\begin{split}
 & \mathrm{L}_{\text{CCL}} = \underset{f}{\rm sup}\,\,\mathbb{E}_{z\sim P_Z} \\ & \left[\mathbb{E}_{(x_i, y_i)\sim {P_{X,Y|z}}}\left[\,\frac{1}{n}\sum_{i=1}^n {\rm log}\,\frac{e^{f(x_i, y_i)}}{\frac{1}{n}\sum_{j=1}^n e^{f(x_i, y_j)}}\right]\right]
\end{split}
\label{eq:ccl}
\vspace{-5pt}
\end{equation}

\noindent where the positive pairs $\{(x_i,y_i)\}_{i=1}^n$ represent samples drawn from the \textit{conditional} joint distribution: $(x_i,y_i) \sim P_{X,Y|z}$, while the negative pairs $\{(x_i,y_{j \neq i})\}$ represent samples drawn from the product of conditional marginal distributions: $(x_i,y_{j \neq i}) \sim P_{X|z}P_{Y|z}$. 
The proposed Conditional Contrastive Learning (CCL) differs from conventional contrastive SSL by selecting positive and negative pairs from distributions conditioned on a sensitive attribute referred to as $Z$. The CCL approach reduces the influence of the sensitive attribute $Z$ by treating $Z$ as the conditional variable between $X$ and $Y$. We focus on scenarios where $Z$ is readily available in the dataset (e.g., gender, race, or age).

\paragraph{Conditional sampling}

We first define a preprocessing stage to identify hard negative samples before the training phase. Each entity $\mathbf{E}_i$ in the knowledge base contains a set of predefined attributes. We calculate the Jaccard distance between two entities $i$ and $j$ as the similarity score between the two entities:

\vspace{-5pt}
\begin{equation}
     J(i,j)=\frac{\left| \mathbf{e}_{a_i} \cap \mathbf{e}_{a_j} \right|}{\left| \mathbf{e}_{a_i} \cup \mathbf{e}_{b_i} \right|}
     \vspace{-7pt}
\end{equation}

For each entity $\mathbf{E}_i$, we calculate the total Jaccard distance with the knowledge base $J_i = \left [ J(i,1), J(i,2), \ldots, J(i,j) \right ]$, where $1 \leqslant j \leqslant n_e$ and $j \neq i$. We store these similarity scores and sort them. Then, we select the top $k$ scores as the $k$ most similar or hard negative samples of $\mathbf{E}_i$: $\hat{J}_i = \text{top}_k , J_i$.

\paragraph{Jaccard Distance-based CCL}

After determining the top $k$ negative samples, we calculate the conditional contrastive learning meta-attribute JD-CCL:

\vspace{-5pt}
\begin{equation}
\begin{split}
& \mathrm{L}_{JD-CCL} = \\
& E_{(\mathbf{M}, \mathbf{E}_{pos})\sim {P_{\mathrm{M},\mathrm{E}}}, z \sim \hat{J}_{\mathbf{E}_{pos}}, \{\mathbf{E}_{neg,j}\}_{j=1}^k \sim P_{E|Z=z_i}} \\ & \left[\,{\rm log}\,\frac{e^{\mathrm{F}(\mathbf{M}, \mathbf{E}_{pos})}}{ e^{\mathrm{F}(\mathbf{M}, \mathbf{E}_{pos})} + \sum_{j=1}^k e^{\mathrm{F}(\mathbf{M}, \mathbf{E}_{neg,j})}}\right]
\end{split}
\vspace{-5pt}
\end{equation}

\noindent where positive pairs $(\mathbf{M}_i, \mathbf{E}_{pos,i})$, consisting of a mention input and its linked entity in the knowledge base, are drawn randomly from the dataset. After sampling a positive pair, negative entities $\{\mathbf{E}_{neg,j}\}_{j=1}^k$ are drawn from the set $\hat{J}i$ of the entity $\mathbf{E}_{pos,i}$. Unlike CCL for fair representation, this approach samples negative entities based on their meta-information, thereby covering different attributes, whereas CCL is based on predefined conditions $Z$ such as gender and race. Furthermore, the difference between JD-CCL and Contrastive Learning can be described as follows: JD-CCL samples negative entities from conditions $z \in Z$, resulting in negative pairs that possess more general features. This increased diversity in features makes the model more challenging to train. 

\subsection{Contextual Visual-aid Controllable Patch Transformation}
\label{sec:cpt}

To address the issue of dissimilarity between visual information in mention inputs and knowledge entities, we propose the Contextual Visual-aid Controllable Patch Transformation (CVaCPT) module. CVaCPT adaptively influences the mention visual input by applying a transformation to each patch representation based on prior information. Technically, we utilize the Stable Diffusion model to generate an image $s_j$ corresponding to the input text $x_j$. Here, $m_j$ is the mention sentence prefixed with "An image of". Thus, the same sentence with different mentions we want to match will create different images of the target. Using a visual encoder $\mathrm{VIS}_{enc}$, we obtain a global representation $VEs_j^G$ of the synthetic image $s_j$.

\paragraph{Controllable patch transformation (CFT)}

The generated image with the target reference text in the sentence helps the model selectively control the important features of the mention input. We propose the Controllable Feature Transformation (CFT) module to manage the information flow from the visual encoder, $VE_j^G$ and $VE_j^L$, before feeding it into the matcher. Specifically, as shown in Fig.~\ref{fig:framework}, the synthetic features $VEs_j^G$ combined with the text mention representation $TE_j^G$ are used to fine-tune the visual features of the mention, $VE_j^G$ and $VE_j^L$, through the transformation of features using the affine parameters $\alpha$ and $\beta$:

\vspace{-5pt}
\begin{align}
\hat{VE}_j^G = VE_j^G +  w(\alpha^G \odot VE_j^G + \beta^G) \\
\hat{VE}_j^L = VE_j^L +  w(\alpha^L \odot VE_j^L + \beta^L) \\
\alpha^G, \beta^G = \mathrm{A}^G( \left [ VE_j^G \cdot VTEs_j \right ] \\
\alpha^L, \beta^L = \mathrm{A}^L( \left [ VE_j^L \cdot VTEs_j \right ] \\
VTEs_j = \mathrm{CE}(\left [VEs_j^G \cdot TE_j^G \right ])
\label{eq:cpt}
\vspace{-5pt}
\end{align}

\noindent where $\mathrm{P}_{\theta^X}$ denotes a stack of convolutions that predicts $\alpha^X$ and $\beta^X$ from the concatenated ($\left[\cdot\right]$) features $\left[ VE_j \cdot \mathrm{CE}(\left[VEs_j^G \cdot TE_j^G \right]) \right]$. $\mathrm{CE}$ is the contextual visual-aid module, which receives a concatenation of the global representation of the synthetic image $VEs_j^G$ and the global representation of the input mention $TE_j^G$. An adjustable coefficient $w \in [0,1]$ is then used to control the relative importance of the inputs. With the CPT module, the mention image representations are adjusted to match the mentioned entity in the sentence, removing noisy and unhelpful patches needed to match in the knowledge base.

\paragraph{Multiview Synthetic image}

We found that images generated from the diffuser model can include noisy features (e.g., background details not related to the mention). Thus, using multiple images can help the CFT module select important and helpful features from synthetic images. We generate $n_s$ different synthetic images from the same prompt $s_{j,h}$, where $h \in \{1, \ldots, n_s\}$. Each global visual representation is concatenated with the global textual representation before being fed into $\mathrm{CE}$:

\vspace{-5pt}
\begin{align}
    \hat{VTEs}_j = \left [  VTEs_{j,1}, ..., VTEs_{j,n_s} \right ] \\
    VTEs_{j,h} = \mathrm{CE}(\left [VEs_{j,h}^G \cdot TE_j^G \right ])
    \vspace{-7pt}
\end{align}

We apply max-pooling to calculate the maximum value for patches of a feature map across multiple images:

\vspace{-5pt}
\begin{equation}
    VTEs_j = \mathrm{max\_pooling} ( \hat{VTEs}_j )
    \vspace{-7pt}
\end{equation}

This $\mathrm{max\_pooling}$ operation emphasizes the presence of strong/important features and diminishes the noise.

\begin{table*}[t]

\begin{center}
\scalebox{0.85}{
\begin{tabular}{l|l|l|l|l|l|l|l|l|l|l|l|l}
\toprule
\multirow{2}{*}{Model} & \multicolumn{4}{c|}{WikiDiverse}                                                                                                                   & \multicolumn{4}{c|}{RichpediaMEL}                                                                                                                  & \multicolumn{4}{c}{WikiMEL}   \\

\cline{2-13} 

                       & H@1                                & H@3                                & H@5                                & MRR                                & H@1                                & H@3                                & H@5                                & MRR                                & H@1   & H@3   & H@5   & MRR   \\ \hline
CLIP                   & 61.21                              & 79.63                              & 85.18                              & 71.69                              & 67.78                              & 85.22                              & 90.04                              & 77.57                              & 83.23 & 92.10 & 94.51 & 88.23 \\
ViLT                   & 34.39                              & 51.07                              & 57.83                              & 45.22                              & 45.85                              & 62.96                              & 69.80                              & 56.63                              & 72.64 & 84.51 & 87.86 & 79.46 \\
ALBEF                  & 60.59                              & 75.59                              & 81.30                              & 69.93                              & 65.17                              & 82.84                              & 88.28                              & 75.29                              & 78.64 & 88.93 & 91.75 & 84.56 \\
METER                  & 53.14                              & 70.93                              & 77.59                              & 63.71                              & 63.96                              & 82.24                              & 87.08                              & 74.15                              & 72.46 & 84.41 & 88.17 & 79.49 \\
MIMIC                  & 63.51                              & 81.04                              & 86.43                              & 73.44                              & 81.02                              & 91.77                              & \textbf{94.38}                              & 86.95                              & 87.98 & 95.07 & 96.37 & 91.82 \\ \hline
Ours                    & \textbf{70.25} & \textbf{83.78} & \textbf{88.06} & \textbf{78.09} & \textbf{83.38} & \textbf{91.91} & 94.18 & \textbf{88.14} & \textbf{89.28} & \textbf{95.31} & \textbf{96.84} & \textbf{92.62} \\ \bottomrule

\end{tabular}}

\caption{Performance comparison on three MEL tasks: WikiDiverse, RichpediaMEL, and WikiMEL. We trained the models using a random seed of $5$, presenting the average of our findings. Our experiments included paired t-tests with $p<0.05$, demonstrating overall significant improvement.}

\label{tab:main}
\end{center}

\vspace{-11pt}
\end{table*}

\section{Experiments setups}

\subsection{Datasets}

In the experiments, we adopt three benchmark MEL datasets:
\textbf{(i) WikiDiverse} \cite{DBLP:conf/acl/WangTGLWYCX22} is a meticulously curated MEL dataset featurthe appendixverse array of contextual topics and entity types sourced from Wikinews. Comprising 8,000 image-caption pairs, it employs Wikipedia as its underlying knowledge base.
\textbf{(ii) WikiMEL} \cite{10.1145/3477495.3531867} comprises over 22,000 multimodal sentences gathered from Wikipedia entity pages. 
\textbf{(iii) RichpediaMEL} \cite{DBLP:journals/bdr/WangWQZ20} Initially, entities were extracted from Richpedia for the creation, followed by the acquisition of multimodal data from Wikidata~\cite{DBLP:journals/cacm/VrandecicK14}. Individuals constitute the primary entity category in both WikiMEL and RichpediaMEL.

\subsection{Baselines}

We compared our method with various competitive VLP models: \textbf{CLIP}~\cite{DBLP:conf/icml/RadfordKHRGASAM21},  \textbf{ViLT}~\cite{DBLP:conf/icml/KimSK21}, \textbf{ALBEF}~\cite{DBLP:conf/nips/LiSGJXH21}, \textbf{METER}~\cite{DBLP:conf/cvpr/DouXGWWWZZYP0022}, \textbf{MIMIC}~\cite{luo2023multigrained}. Detailed descriptions are provided in the Appendix~\ref{apx:baselines}.

\subsection{Metrics}

We calculated the similarity between a mention and all entities in the knowledge base (KB) to measure their aligning probability. The similarity scores are sorted in descending order to calculate \textbf{H@1}, \textbf{H@3}, \textbf{H@5}, and \textbf{MRR}. \textbf{H@k} indicates the hit rate of the ground truth entity when only considering the top-k ranked entities: \textbf{H@1,3,5} represent the hit rates in the top 1, 3, and 5, respectively. \textbf{MRR} represents the mean reciprocal rank of the ground truth entity: $\frac{1}{N}\sum_{i=1}^{N}\frac{1}{rank(i)}$.

\subsection{Implementations}

We initialize the CLIP model (Vit-Base-Patch32) as both our visual encoder $\mathrm{VIS}_{enc}$ and text encoder $\mathrm{TEXT}_{enc}$, from HuggingFace.
For visual input, all images (mention, knowledge entity and synthetic image) are rescaled to 224 $\times$ 224. With the mention or entity input that do not have image, will initialize with blank image (while) with zero padding. For textual input, we use CLIP tokenizer, which constructs based on byte-level Byte-Pair Encoding. We implement linear neural networks to project the hidden representation from CLIP model to $96$ and setup the matcher similar to MIMIC. Since our CVaCPT module and contrastive objective are flexibly integrated with various models, we built upon the implementation of the state-of-the-art MIMIC~\cite{luo2023multigrained} by incorporating CVaCPT and replacing the original contrastive learning objective with our JD-CCL approach.

\begin{table}[t]

\begin{center}
\scalebox{0.85}{
\begin{tabular}{c|c|c|c|c}
\toprule
Setting           & H@1        & H@3       & H@5       & MRR      \\
\midrule
Our             & 70.25      & 83.78     & 88.06     & 78.09    \\
w/o LD-CCL      & 66.55      & 82.57      & 87.53      & 75.72     \\
w/o CVaCPT      & 67.80       & 81.85      & 86.62      & 76.04     \\
\midrule
\multicolumn{5}{c}{(1) Number of negative samples in LD-CCL}        \\
\midrule
2               & 67.88      & 82.57     & 87.24     & 77.26    \\
4               & 69.00      & 83.01     & 87.34     & 77.08    \\
6               & 69.53      & 84.16     & 88.30     & 77.80    \\
\midrule
\multicolumn{5}{c}{(2) Number of synthetic images in CVaCPT module} \\
\midrule
1               & 66.68      & 82.85     & 87.10     & 76.04    \\
2               & 68.47      & 83.78     & 88.16     & 77.03    \\
3               & 68.67      & 83.54     & 86.86     & 76.87   \\
\bottomrule
\end{tabular}
}
\caption{Ablation results on WikiDiverse.}

\label{tab:abl1}
\end{center}

\vspace{-12pt}
\end{table}

\begin{table}[t]

\begin{center}
\scalebox{0.85}{
\begin{tabular}{l|c|c|c|c}
\toprule
Model & H@1            & H@3            & H@5            & MRR            \\
\midrule
\multicolumn{5}{c}{WikiDM}                                                \\
\midrule
MIMIC & 74.77          & 89.12          & 92.25          & 82.63          \\
Our   & \textbf{78.77} & \textbf{90.25} & \textbf{92.92} & \textbf{85.16} \\
\midrule
\multicolumn{5}{c}{WikiDR}                                                \\
\midrule
MIMIC & 72.41          & 84.75          & 89.00          & 79.63          \\
Our   & \textbf{77.32} & \textbf{88.65} & \textbf{91.70} & \textbf{83.59} \\ \bottomrule

\end{tabular}
}
\caption{Ablation results WikiDM and WikiDR.}

\label{tab:abl2}
\end{center}

\vspace{-12pt}
\end{table}

\section{Experimental results}

Following the methodology of previous studies, we conducted our model five times using the same hyperparameter settings. The average results are presented in Table~\ref{tab:main}. An overview of our results reveals that our model consistently matches or surpasses the benchmarks set by baseline approaches. Our proposed method shows significant improvements across various datasets. Notably, we achieved a substantial increase of $6.74\%$ in accuracy for H@1 on WikiDiverse, and improvements of $2.36\%$ and $1.30\%$ on RipediaMEL and WikiMEL, respectively. For H@3, H@5, and MRR, we also achieved better performance on most datasets. To be more specific, we outperform the state-of-the-art (SOTA) by $2.74\%$, $1.63\%$, and $4.65\%$ on WikiDiverse, and achieve slight improvements on the other two datasets, except for H@5 metric on WikiMEL.
\begin{figure}[t]
     \centering
     \begin{subfigure}[b]{0.49\textwidth}
         \centering
         \includegraphics[width=0.90\textwidth]{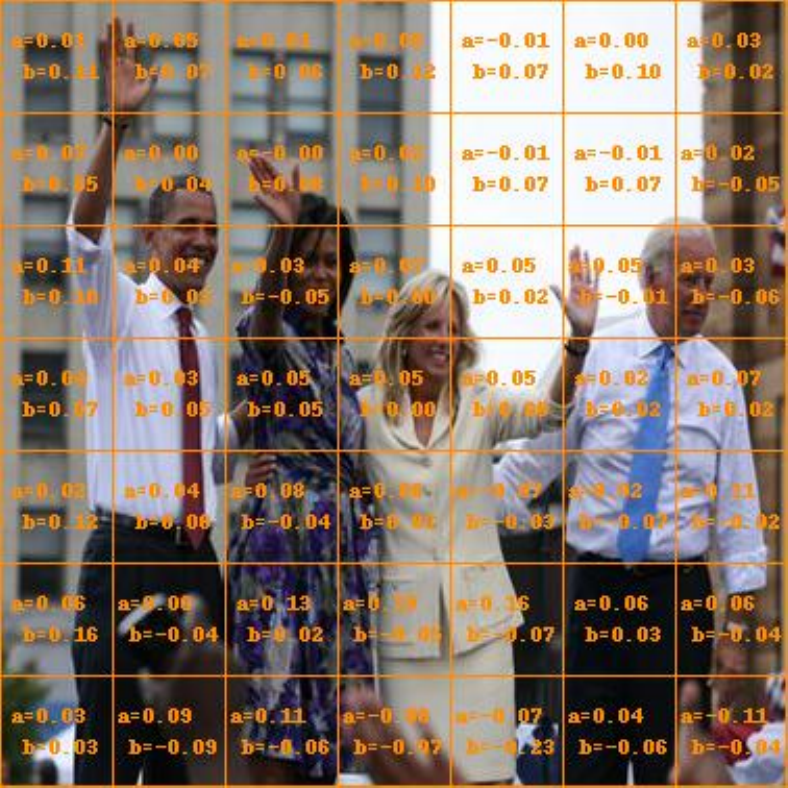}
         \caption{Sentence: Barack Obama, Michelle Obama, Jill Biden and Joe Biden at the United States Vice Presidential announcement on August 23, 2008. Mention: Barack Obama.}
         \label{fig:vis1}
     \end{subfigure}
     \hfill
     \begin{subfigure}[b]{0.49\textwidth}
         \centering
         \includegraphics[width=0.90\textwidth]{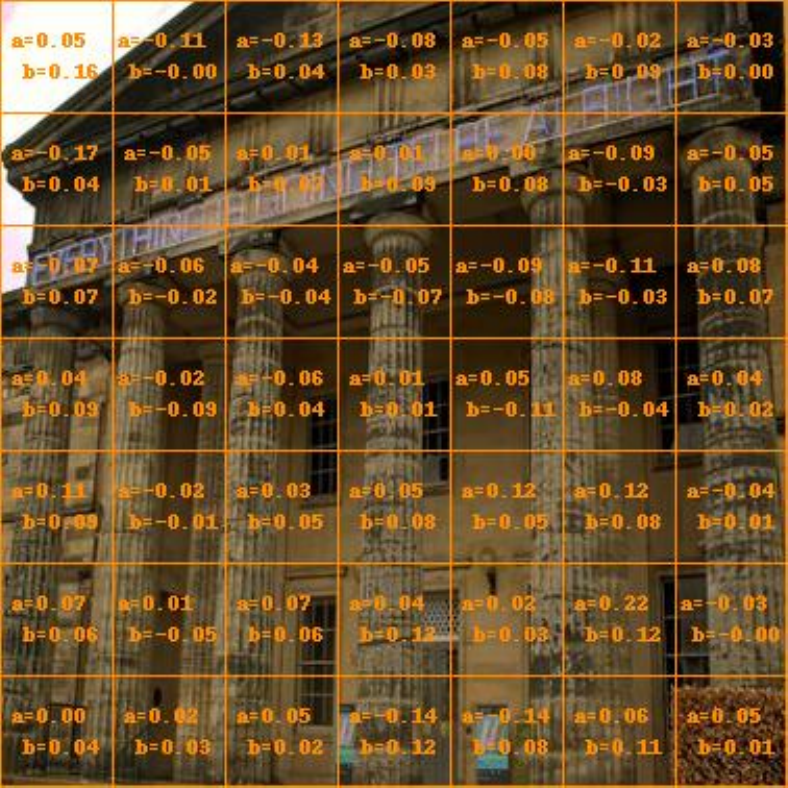}
         \caption{Sentence: The exterior of the Modern One building of Edinburgh's Scottish National Gallery of Modern Art. Mention: Modern One.}
         \label{fig:vis2}
     \end{subfigure}

     \caption{Visualization of CVaCPT module}

     \vspace{-13pt}
\end{figure}

\section{Analysis}

\subsection{Regularization}

To validate the significance of each of our objectives, we conducted additional experiments where one of our proposed methods was removed. As illustrated in Table~\ref{tab:abl1}, the results are presented in the first three rows, indicating that incorporating all the proposed approaches leads to the best performance. The second row shows that not applying JD conditional contrastive learning harms performance. This result can be attributed to the fact that, during training, hard negative samples play an important role in this task. In the third row of the table, we demonstrate that the removal of CVaCPT significantly affects performance across all metrics.

\subsection{Number of negative samples and number of synthetic images}

We conducted an experiment to better understand the impact of the number of negative samples, $k$, in JD-CCL and the number of synthetic images, $s$, in CVaCPT. We set $k$ to values in the set ${2, 4, 6}$ and $s$ to values in the set ${1, 2, 3}$. As shown in the first three rows of Table~\ref{tab:abl1}, a larger number of conditional negative samples ($k=4, 6$) achieves better results compared to a smaller number ($k=2$), although there is only a small difference between $k=4$ and $k=6$. Regarding the number of synthetic images, using $s=1$ means not applying the max-pooling operation. As shown in the last three rows, multiple views of visual augmentation outperform a single view ($s=1$) with only one synthetic image. However, increasing the number of images from 2 to 3 does not significantly improve the model.

\subsection{Larger knowledge base evaluation}

While the benchmarks WikiMEL and RichpediaMEL achieved good performance, WikiDiverse appears to be a more challenging dataset. To enhance the ability to handle a larger knowledge base, we merged the two easier benchmarks with the more challenging WikiDiverse to create two larger datasets (double in size): WikiDM (WikiDiverse + WikiMEL) and WikiDR (WikiDiverse + RichpediaMEL). The experimental results are presented in Table~\ref{tab:abl2}. As shown, our approach outperforms MIMIC across all metrics on both datasets. Specifically, our method achieves an improvement of $4.00\%$ and $4.91\%$ in the H@1 metric on WikiDM and WikiDR, respectively, and outperforms in all other metrics as well.

\subsection{Visualization}

In this appendix, we provide a visualization of the Controllable Patch Transformation module, as illustrated in Figures~\ref{fig:vis1} and \ref{fig:vis2}. We visualize the calculation of alpha and beta for patches in the image corresponding to the input mention (in each caption) and synthetic images. As shown in Figure~\ref{fig:vis1}, the model focuses on people in the image (alpha and beta values are large) while reducing the impact of the background (sky, building) (alpha and beta values are small). In contrast, Figure~\ref{fig:vis2} shows a building covering the entire image, where all alpha and beta values are equally high, indicating that all image patches are informative for the matcher.

\section{Conclusion}

This paper introduces JD-CCL, a novel conditional contrastive learning approach based on Jaccard Distance. This method leverages meta-attributes to draw conditional negative samples with similar features during training, allowing the model to identify specific characteristics for prediction accurately. Additionally, our proposed Contextual Visual-Aid Controllable Patch Transform (CVa-CPT) module enhances visual modality representation by utilizing synthetic images and contextual information from the mention sentence. Our framework paves the way for further exploration in this domain, promoting multimodal and knowledge base application advancements.
\section*{Limitations}

Despite the improvements our approach achieves in the MEL task, JD-CCL and CVaCPT still have the following limitations. CVaCPT uses a Text-to-Image diffuser model to generate synthetic images, which requires additional time and computational resources. Furthermore, using synthetic images from a different domain might introduce some noise due to the domain gap. Secondly, JD-CCL includes a condition sampling stage, which is a preprocessing step to determine the negative samples, and this stage also takes time. Additionally, we automatically calculate the Jaccard distance between the attribute sets of two entities. If the dataset is not well-annotated, this could negatively impact the training stage.


\bibliography{custom}

\appendix
\appendix
\onecolumn
\section{Baselines}
\label{apx:baselines}

We followed the baseline setup from ~\cite{luo2023multigrained}:

\paragraph{CLIP} leverages two Transformer-based encoders to generate visual and textual representations, pre-training on vast amounts of noisy web data using contrastive loss between 2 global representations of each modality.

\paragraph{METER} adopts a co-attention schema to explore the semantic relationships between different modalities, with each layer comprising a self-attention module, a cross-attention module, and a feed-forward network.

\paragraph{ViLT} applies deep modality interaction through a stack of Transformer layers and proposes utilizing early, shallow textual and visual embeddings.

\paragraph{ALBEF} initially aligns visual and textual features through image-text contrastive loss and then integrates them using a multimodal Transformer encoder. It further employs momentum distillation to enhance learning from noisy data.

\paragraph{MIMIC} represents the current state-of-the-art in MEL tasks across three datasets. MIMIC introduces three modules: the Text-based Global-Local Interaction Unit (TGLU), the Vision-based Dual Interaction Unit (VDLU), and the Cross-Modal Fusion-based Interaction Unit (CMFU), which are designed to capture and integrate fine-grained representations in abbreviated text and implicit visual cues.

\begin{table}[t]

\begin{center}

\begin{tabular}{l|c|c}
CLIP             & Normal feature & Pooling feature \\ \hline
Mention images   & 0.406          & \textbf{0.430}  \\
KB entity images & 0.552          & \textbf{0.585}  \\ \hline
DinoV2           & Normal feature & Pooling feature \\ \hline
Mention images   & 0.178          & \textbf{0.196}  \\
KB entity images & 0.310          & \textbf{0.344} 
\end{tabular}

\caption{Comparison of similarity scores between the mean of individual images and pooling features with KB entity images.}

\label{ap:res}
\end{center}
\end{table}

\begin{table}[]

\begin{center}
\begin{tabular}{c|c|c|c|c|c}
Mention image & KB entity image & \multicolumn{3}{c|}{Synthetic images} & Pooling feature \\ \hline

\multicolumn{6}{c}{\begin{tabular}[c]{@{}l@{}}Mention: \textbf{President Trump} holds a Bible in front of St. John's Episcopal Church.\\ KB entity: President of the United States from 2017 to 2021.
\end{tabular}} \\ \hline

\includegraphics[width=1.8cm,height=1.8cm]{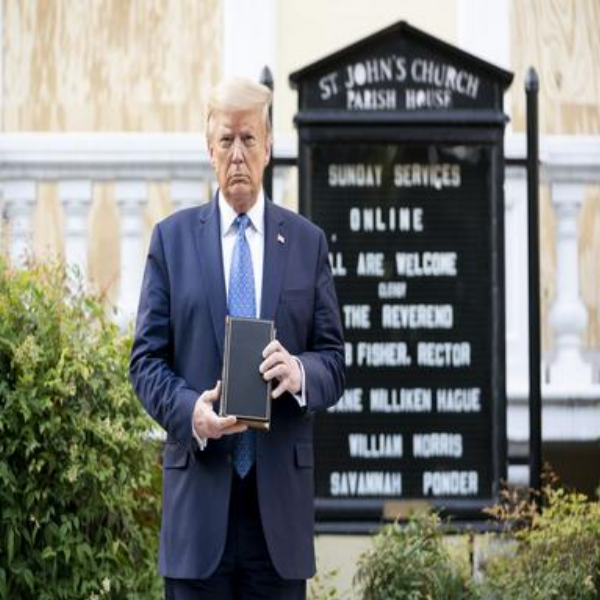} & \includegraphics[width=1.8cm,height=1.8cm]{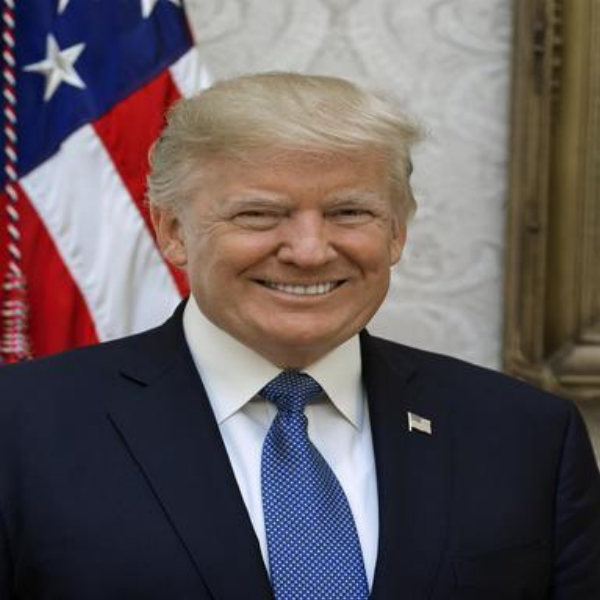} & \includegraphics[width=1.8cm,height=1.8cm]{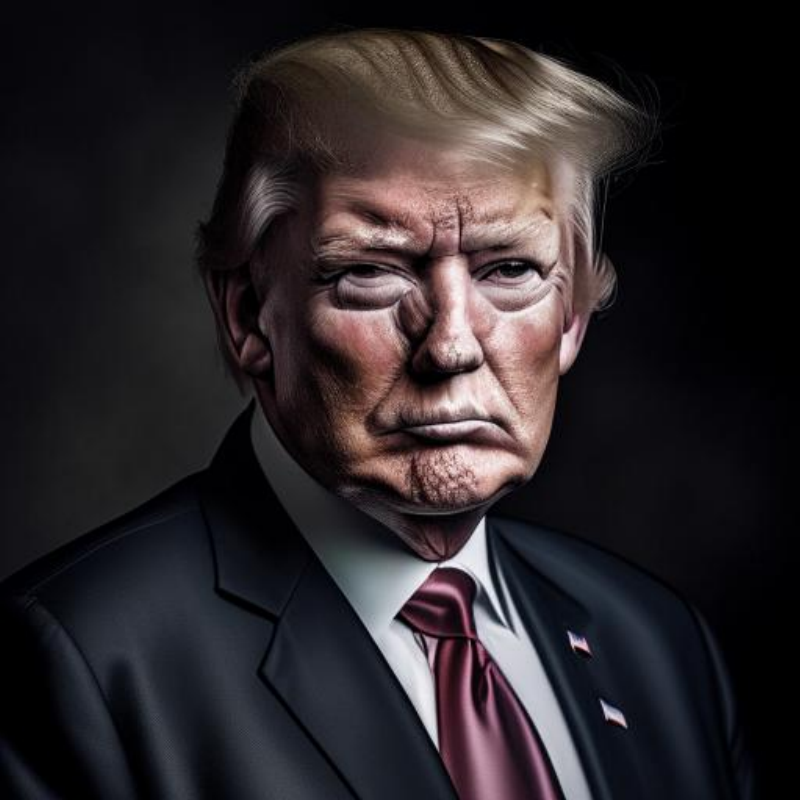} & \includegraphics[width=1.8cm,height=1.8cm]{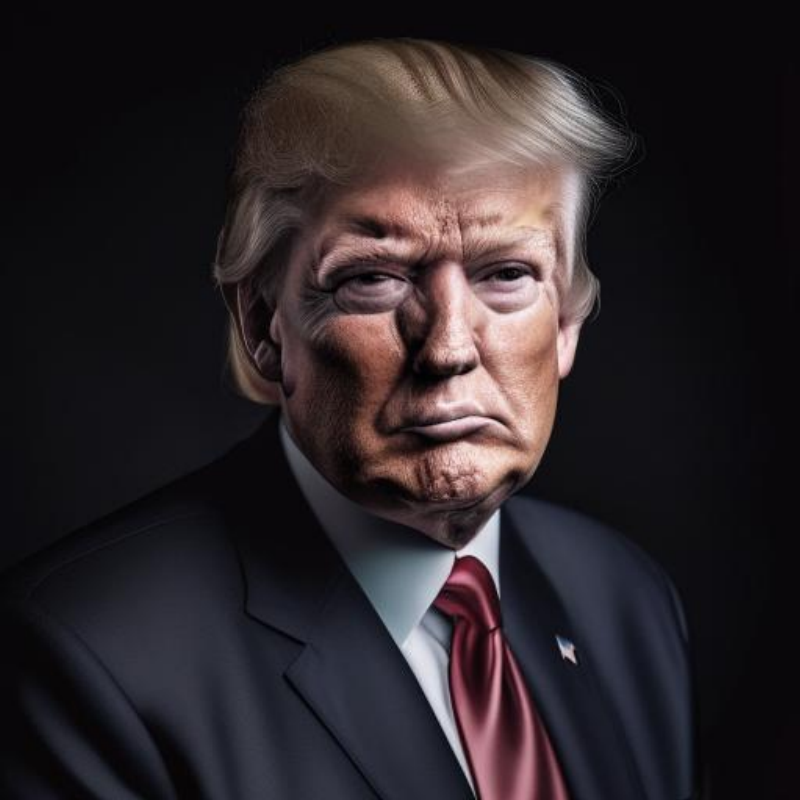} & \includegraphics[width=1.8cm,height=1.8cm]{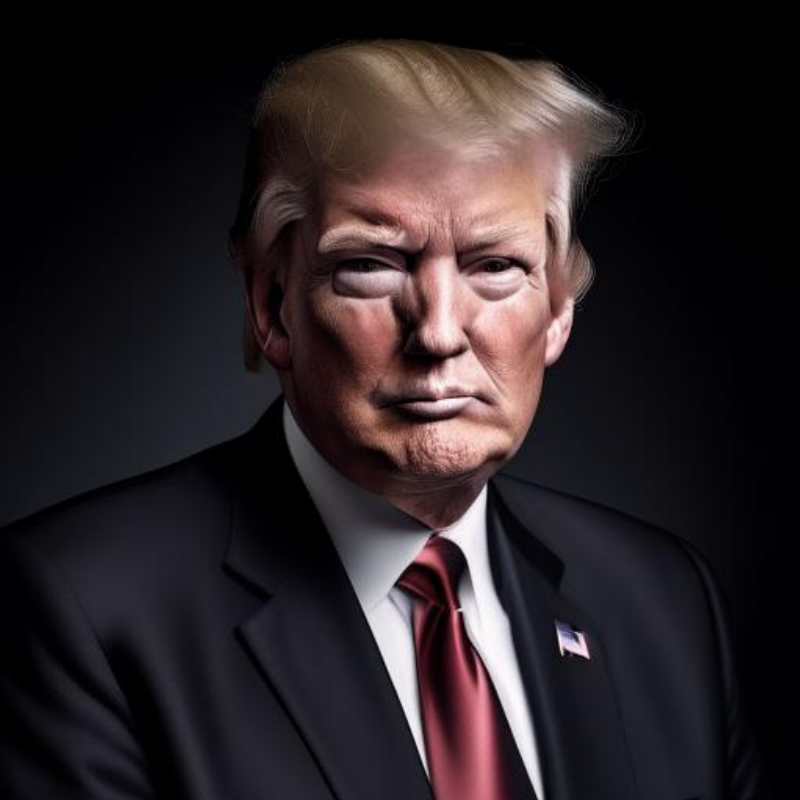}  & 
\includegraphics[width=1.8cm,height=1.8cm]{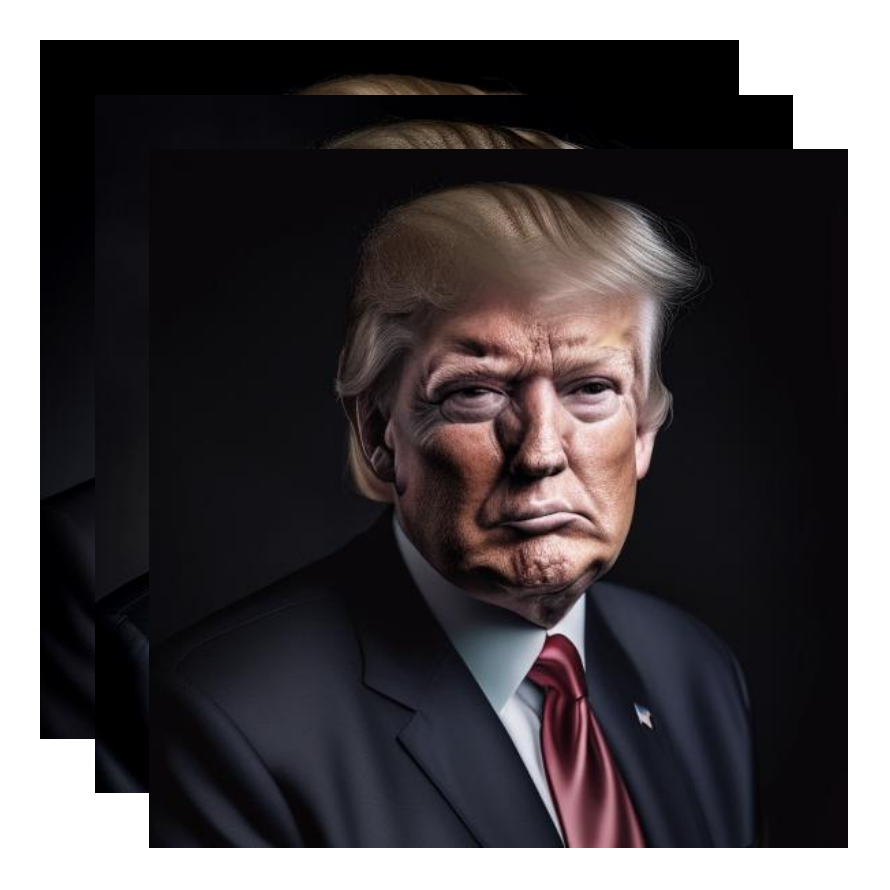}\\ \hline
\multicolumn{2}{c|}{Dino}     & 0.287    & 0.306   & 0.334    & \textbf{0.393}      \\
\multicolumn{2}{c|}{CLIP}     & 0.440    & 0.435    & 0.461    & \textbf{0.518}      \\ \hline

\multicolumn{6}{c}{\begin{tabular}[c]{@{}l@{}}Mention: \textbf{Radiotherapy} and chemotherapy act together to damage the DNA  of cancer \\ cells. According to this new study, gemcitabine potentiates this effect.
\\ KB entity: Therapy using ionizing radiation\end{tabular}} \\ \hline

\includegraphics[width=1.8cm,height=1.8cm]{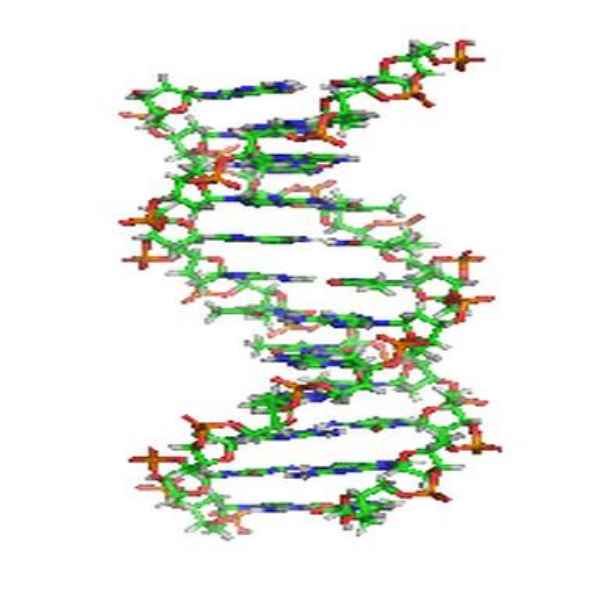}  & \includegraphics[width=1.8cm,height=1.8cm]{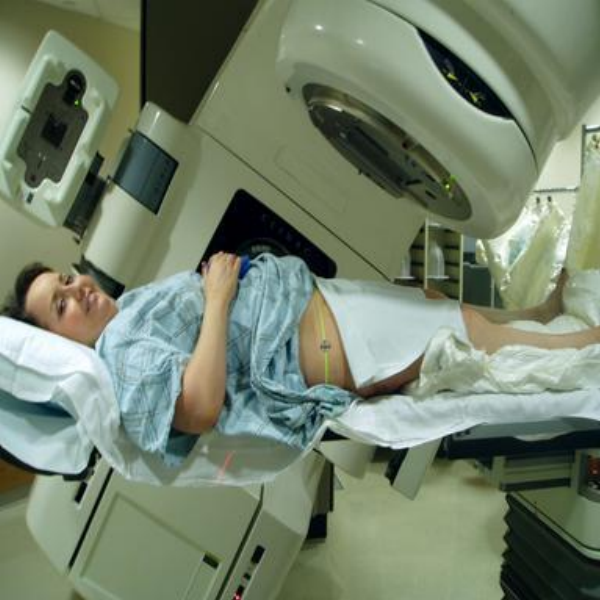} & \includegraphics[width=1.8cm,height=1.8cm]{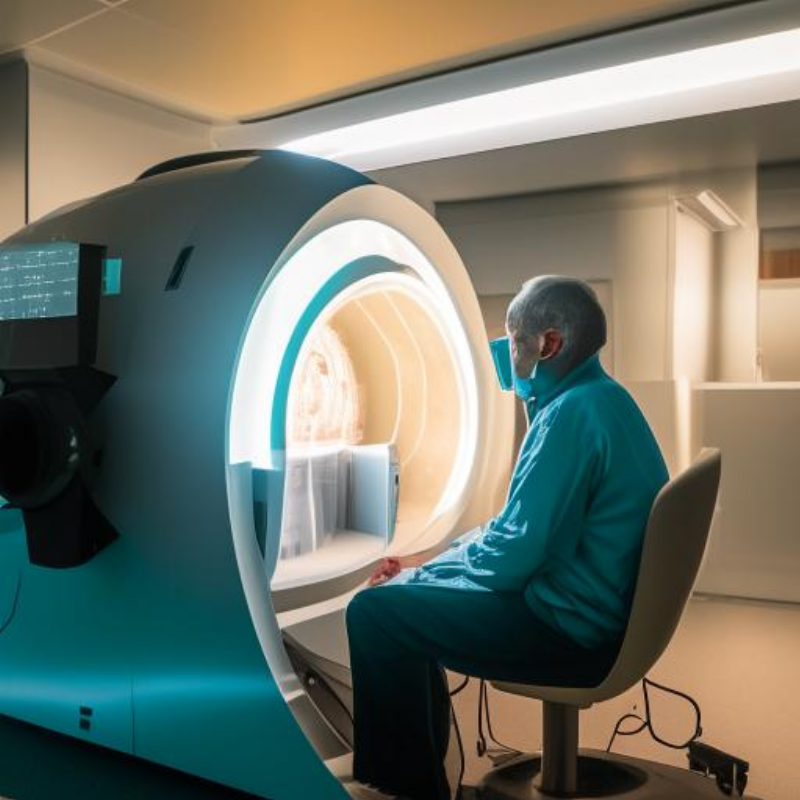} & \includegraphics[width=1.8cm,height=1.8cm]{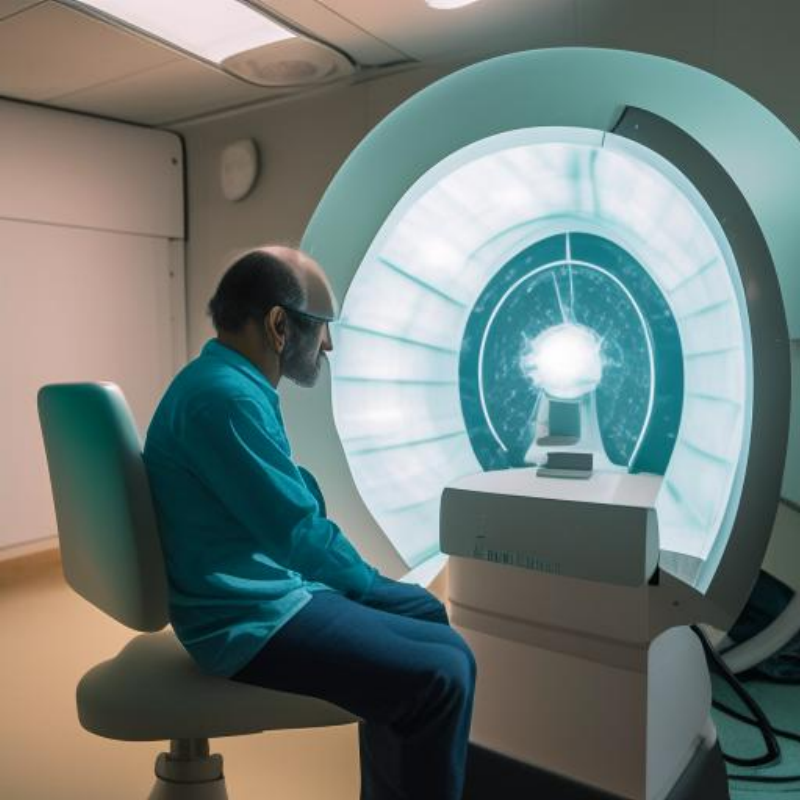} & \includegraphics[width=1.8cm,height=1.8cm]{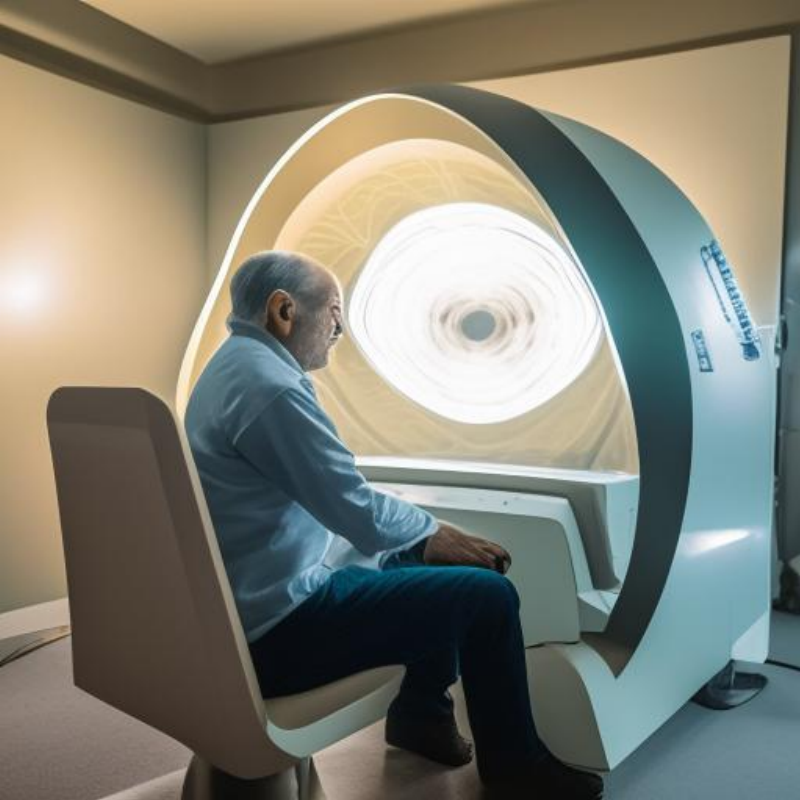} & 
\includegraphics[width=1.8cm,height=1.8cm]{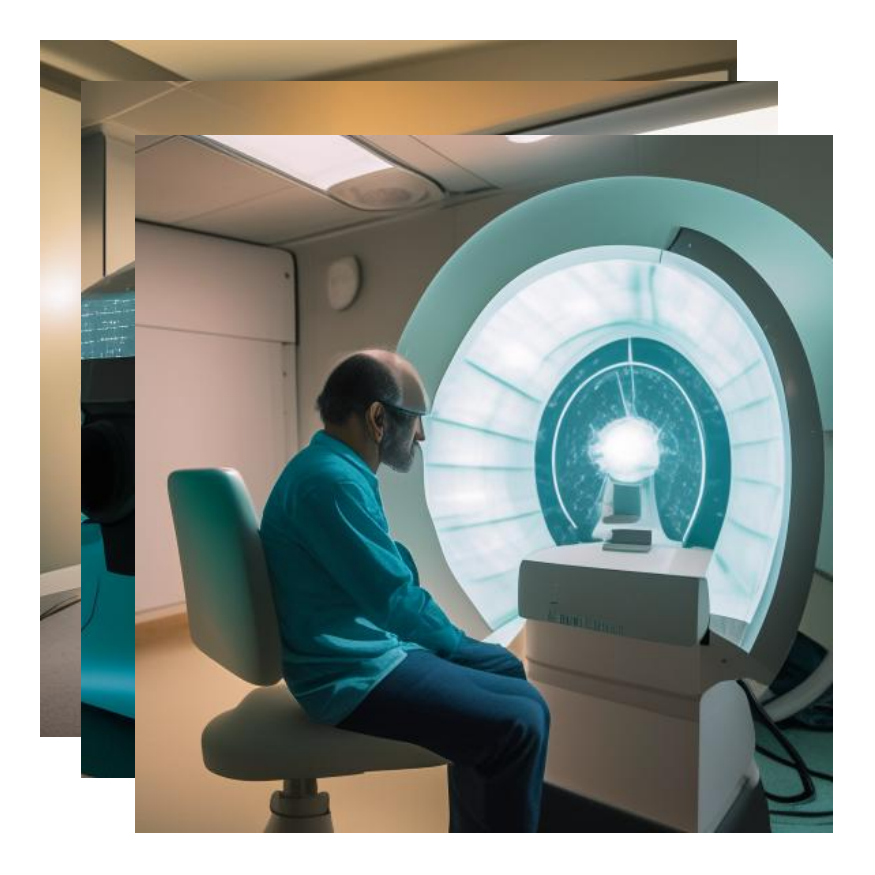}\\ \hline
\multicolumn{2}{c|}{Dino}     & 0.381   & 0.360   & 0.433    & \textbf{0.457}      \\
\multicolumn{2}{c|}{CLIP}     & 0.432   & 0.519    & 0.619    & \textbf{0.631}      \\ \hline
\end{tabular}

\caption{Visualization of synthetic images in the Contextual Visual-aid Controllable Patch Transformation module.}

\label{ab:vis3}
\end{center}

\end{table}

\section{Noisy synthetic images}


We conducted additional experiments to gain further insights into the behavior of the diffusion model in CVaCPT. For each sample in the dataset, we first used the visual encoder to extract features from the mention input image, the knowledge base entity image, and the synthetic image. We then calculated the cosine similarity between each synthetic image and the pooled synthetic image, comparing them to the mention and knowledge base entity images. The mean similarity scores are reported in Table~\ref{ap:res}. As shown, the similarity scores for pooled visual features are higher than those obtained from independent images. This indicates that using multiple synthetic images with a pooling operation helps mitigate the noise introduced by the diffusion model. We also visualize a selection of samples from the datasets in Table~\ref{ab:vis3}, illustrating the diversity of images generated by the model. The similarity scores exhibit a wide distribution across individual images, while the pooled features consistently provide better alignment.

\begin{table}[t]

\begin{center}
\begin{tabular}{l|c|c|c|c}
\textbf{} & \textbf{(1)} & \textbf{(2)} & \textbf{(3)} & \textbf{(4)} \\ \hline
\multicolumn{5}{c}{Wikidiverse}                                       \\ \hline
Train     & 8752         & 2517         & 66           & 16           \\
Valid     & 1275         & 384          & 5            & 0            \\
Test      & 1599         & 461          & 15           & 3            \\ \hline
\multicolumn{5}{c}{RichpediaMEL}                                      \\ \hline
Train     & 10992        & 86           & 1375         & 10           \\
Valid     & 3194         & 19           & 348          & 1            \\
Test      & 1545         & 16           & 219          & 0            \\ \hline
\multicolumn{5}{c}{WikiMEL}                                           \\ \hline
Train     & 17725        & 367          & 0            & 0            \\
Valid     & 2543         & 42           & 0            & 0            \\
Test      & 5066         & 103          & 0            & 0           
\end{tabular}

\caption{(1) Both mention and knowledge base have the image. (2) Mention has an image but its knowledge base does not have an image. (3) Mention does not have an image but its knowledge base has an image. (4) Both the mention and knowledge base do not have the image.
}

\label{ab:stat}
\end{center}

\end{table}

\begin{table}[t]

\begin{center}

\scalebox{0.9}{
\begin{tabular}{l|c|c|c|c|c}
           & \small Mention input & \small KB entity & \multicolumn{3}{c}{\small Other KBs}  \\ \hline

            & & & & & \\
           
           \small Image & \includegraphics[width=1.8cm,height=1.8cm]{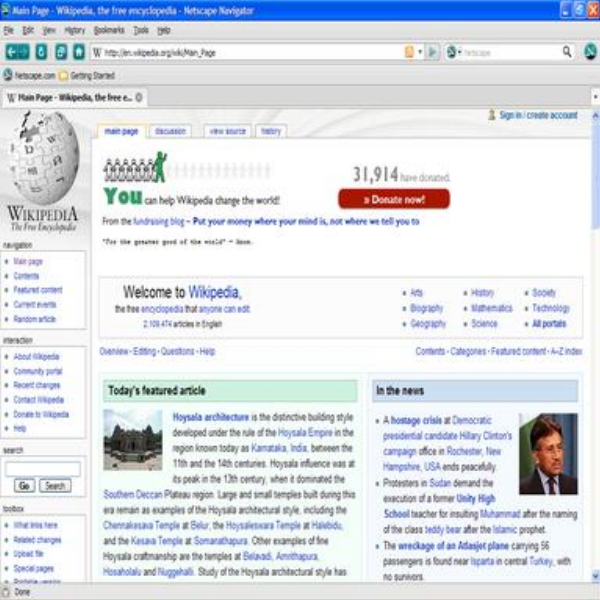} & \includegraphics[width=1.8cm,height=1.8cm]{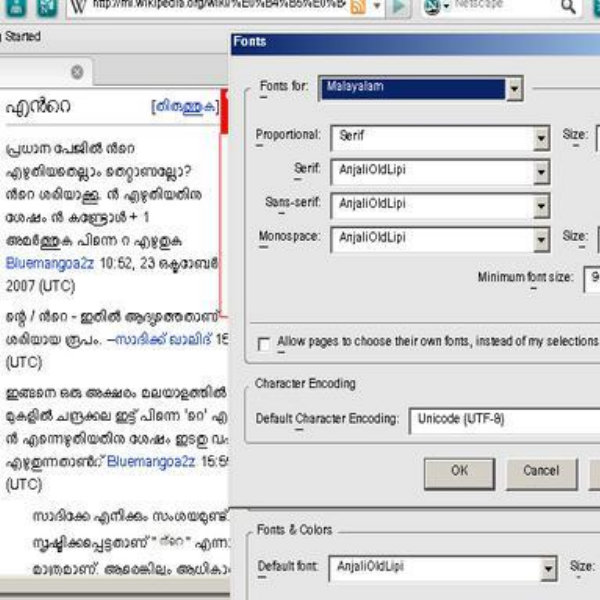} & \includegraphics[width=1.8cm,height=1.8cm]{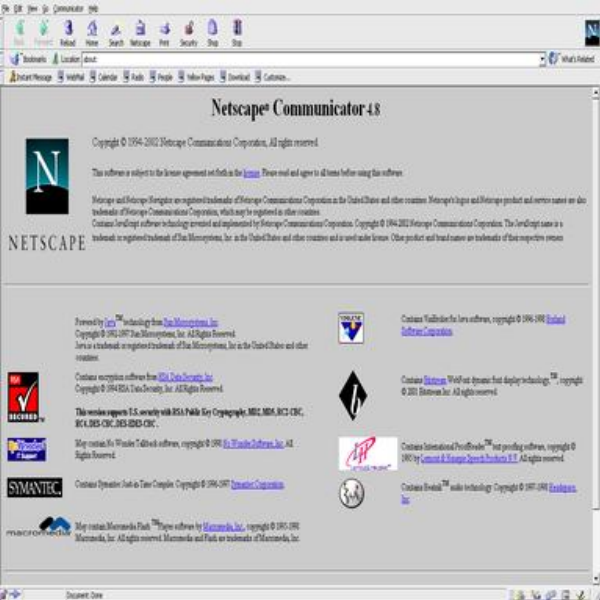} & \small Missing & \small Missing \\ \hline
           
           \small Description & \scalebox{0.85}{\begin{tabular}[c]{@{}l@{}}The Wikipedia home \\ page as viewed on a \\ \textbf{Netscape} browser.\end{tabular}}
            & \scalebox{0.85}{\begin{tabular}[c]{@{}l@{}}Netscape - American \\ computer services \\ company.\end{tabular}}
             & \scalebox{0.85}{\begin{tabular}[c]{@{}l@{}}Netscape Communi\\ cator  - Discontinued  \\Internet  software \\suite\end{tabular}}
             & \scalebox{0.85}{\begin{tabular}[c]{@{}l@{}}Netscape - family \\ of web browsers\end{tabular}}
             & \scalebox{0.85}{\begin{tabular}[c]{@{}l@{}}Netscape 6 - \\version 6 of the \\Netscape web \\browser\end{tabular} }         \\ \hline
\small MIMIC      &        & 3       & 1       & 2         & 4          \\
\small Our method &        & 3      & 1       & 2        & 4          \\ \hline
           Image & \includegraphics[width=1.8cm,height=1.8cm]{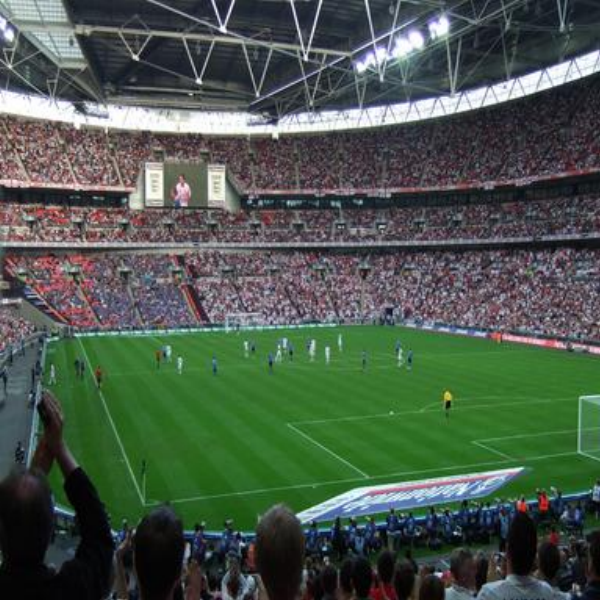} & \small Missing & \small Missing & \small Missing & \small Missing \\ \hline
           Description & \scalebox{0.85}{\begin{tabular}[c]{@{}l@{}}England did win this\\ October 13, 2007 match\\ versus \textbf{Estonia}, 3-0 at\\ Wembley Stadium
\end{tabular}}
           & \scalebox{0.85}{\begin{tabular}[c]{@{}l@{}}Estonia sovereign \\state in northeastern \\Europe\end{tabular}}
           & \scalebox{0.85}{\begin{tabular}[c]{@{}l@{}}Estonia men's \\national basketball \\team\end{tabular}}
           & \scalebox{0.85}{\begin{tabular}[c]{@{}l@{}}Estonia national \\football team \\results\end{tabular}}
           & \scalebox{0.85}{\begin{tabular}[c]{@{}l@{}}Estonia national \\football team\end{tabular}}
           \\ \hline
\small MIMIC      &        & 4       & 3       & 2         & 1          \\
\small Our method &        & 3       & 4       & 1         & 2          \\ \hline

\end{tabular}}

\caption{Visualization of failed cases, samples taken from WikiDiverse}

\label{ab:vis2}
\end{center}
\end{table}

\section{H\@3 and H\@5 insights}

Our method demonstrated improved performance on MEL benchmarks compared to baselines. However, the gains in H\@3 and H\@5 metrics were marginal. To investigate this issue, we conducted additional experiments and data analysis. First, we trained and evaluated our method alongside MIMIC under identical settings, revealing that both models faced the same challenges when tested on the same samples in the test set. Upon further examination, we observed that while the correct entities were often ranked within the top 3 or 5, these mentions were consistently mislinked to entities with the same name in the knowledge base (i.e., same name but representing different entities). Table~\ref{ab:vis2} illustrates two randomly selected failure cases, highlighting several factors that contributed to incorrect predictions: (1) the entities involved shared identical names, (2) these entities were never linked to any mention in the training set, and (3) many of the input entities or mentions lacked corresponding images. For issue (1), we provide a list of duplicate entity names along with their frequency  larger than 10 in the knowledge base in Sec~\ref{ab:similar}. Regarding issue (2), the sheer size of the knowledge base relative to the inputs of mention in the data set resulted in numerous entities, particularly those with duplicate names, being excluded from the training process. Finally, as shown in Table~\ref{ab:stat}, a portion of entity-mention pairs across the three datasets used in this study lacked images for at least one of the mentions or knowledge base entities. These challenges, (1), (2), and (3), collectively hinder the model's ability to fully optimize for such cases.

\section{Similar knowledge entities}
\label{ab:similar}
{\fontfamily{qcr}\selectfont
Washington County (19), John Murphy (18), Greg Smith (18), Black Friday (17), St. John's Episcopal Church (17), Vanity Fair (17), Oliver Twist (17), Washington Township (17), Monroe County (17), Union County (17), Inside Out (16), Masquerade (16), Steve Smith (16), Lockdown (16), Wayne County (16), Montgomery County (16), Alice in Wonderland (15), Red Line (14), House of Representatives (14), Mount Vernon (14), Woodstock (13), Gordon Brown (13), Moby Dick (13), Paul Ryan (13), Government House (13), John Baird (13), Burlington (12), John Edwards (12), La Grange (12), Victoria Park (12), Lake County (12), Stephen Smith (12), Grafton (12), Michael Gallagher (12), Peter Marshall (12), Scarecrow (12), Marshall County (12), Independence Day (12), Greenfield (12), Polk County (12), Ceres (11), Arcadia (11), Homecoming (11), Camera Obscura (11), Equinox (11), Jack Thompson (11), Black Widow (11), San Fernando (11), Scott Brown (11), Supernova (11), White City (11), William Johnson (11), Brian Moore (11), Brush Creek (11), Chief of the Defence Staff (11), Calhoun County (11), Fayette County (11).

}

\end{document}